\documentclass{article}

\usepackage{times}
\usepackage{graphicx} % more modern
\usepackage{natbib}

\usepackage{algorithm}
\usepackage{algorithmic}

\usepackage{wrapfig}

\usepackage[nohyperref,accepted]{icml2017}

\usepackage{url}
\usepackage{amsmath}
\usepackage{bm}
\usepackage{amsthm}
\usepackage{amssymb}
\usepackage{amsfonts,eucal,amsbsy}
\usepackage{mathtools}
\usepackage{booktabs}
\usepackage{caption}
\usepackage{subcaption}
\usepackage{multirow}
\usepackage{wrapfig}
\usepackage{xcolor, colortbl}
\usepackage{fixltx2e}
\usepackage{pbox}
\usepackage{setspace}

\DeclareMathOperator{\E}{\mathbb{E}}

\icmltitlerunning{Improved Variational Autoencoders for Text Modeling using Dilated Convolutions}

\begin{document}

\twocolumn[
\icmltitle{Improved Variational Autoencoders for Text Modeling using Dilated Convolutions}



% you can specify symbols, otherwise they are numbered in order

\begin{icmlauthorlist}
\icmlauthor{Zichao Yang}{cmu}
\icmlauthor{Zhiting Hu}{cmu}
\icmlauthor{Ruslan Salakhutdinov}{cmu}
\icmlauthor{Taylor Berg-Kirkpatrick}{cmu}
\end{icmlauthorlist}

\icmlaffiliation{cmu}{Carnegie Mellon University}

\icmlcorrespondingauthor{Zichao Yang}{zichaoy@cs.cmu.edu}
% \icmlcorrespondingauthor{Zhiting Hu}{zhitingh@cs.cmu.edu}
% \icmlcorrespondingauthor{Ruslan Salakhutdinov}{rsalakhu@cs.cmu.edu}
% \icmlcorrespondingauthor{Taylor Berg-Kirkpatrick}{tbergkir@cs.cmu.edu}

\icmlkeywords{boring formatting information, machine learning, ICML}

\vskip 0.3in
]



\printAffiliationsAndNotice{}  % leave blank if no need to mention equal contribution
% \printAffiliationsAndNotice{} % otherwise use the standard text.
%\footnotetext{hi}

\begin{abstract}
Recent work on generative text modeling has found that variational
autoencoders (VAE) with LSTM decoders perform worse than
simpler LSTM language models \citep{bowman2015generating}. This negative
result is so far poorly understood, but has been attributed to the
propensity of LSTM decoders to ignore conditioning information from the encoder. In this paper, we experiment with a new type of decoder for VAE: a dilated CNN.
By changing the decoder's dilation architecture, we control the size of context
from previously generated words. In experiments, we find that there is a trade-off
between contextual capacity of the decoder and effective use of encoding information. We show that when carefully managed, VAEs can outperform LSTM language models. We demonstrate perplexity gains on two datasets, representing the first positive language modeling result with VAE.
% Further, we conduct an in-depth investigation of the use of VAE (with our new decoding architecture) for semi-supervised tasks, demonstrating gains over several strong baselines.
Further, we conduct an in-depth investigation of the use of VAE
(with our new decoding architecture) for semi-supervised and unsupervised
labeling tasks, demonstrating gains over several strong baselines.
%With the resulting models,
%  we are able generate text conditioning on label, i.e., generate text conditioning
%  on topic and generate review based on rating.
\end{abstract}

\section{Introduction}
Generative models play an important role in NLP, both in their use as language models and because of their ability to effectively learn from unlabeled data.
By parameterzing generative models using neural nets, recent work has proposed model
classes that are particularly expressive and can pontentially model a wide range of phenomena in language and other modalities. We focus on a specific
instance of this class: the variational autoencoder\footnote{The name VAE is often used to refer to both a model class and an associated inference procedure.} (VAE)~\cite{kingma2013auto}.
% inference procedure.

% a model whose common name conflates structure with
% inference procedure:
% Markov chain Monte Carlo sampling~\cite{andrieu2003introduction}
% and variational inference~\cite{beal2003variational} are typically used to approximate
% the posterior distribution in learning generative models.

The generative story behind the VAE (to be described in detail in the next section)
is simple: First, a continuous latent representation is sampled from a multivariate Gaussian. Then, an output is sampled from a distribution parameterized by a neural decoder, conditioned on the
latent representation. The latent representation (treated as a latent variable during training) is intended to give the model more expressive capacity when compared with simpler
neural generative models--for example, conditional language models. The choice of decoding architecture and final output distribution, which connect the latent representation to output, depends on the kind of data being modeled. The VAE owes its name to an accompanying variational technique ~\cite{kingma2013auto} that has been successfully used to train such models on image data ~\cite{gregor2015draw,salimans2015markov,yan2016attribute2image}.
%when paired with the appropriate decoder (for vision, typically an XXX).

The application of VAEs to text data has been far less\ successful~\cite{bowman2015generating,miao2016neural}. The obvious choice for decoding architecture for a textual VAE is an LSTM, a typical workhorse in NLP. However, \citet{bowman2015generating} found that using an LSTM-VAE for text modeling yields higher perplexity on held-out data than using an LSTM language model. In particular, they observe that the LSTM decoder in VAE does not make effective use of the latent representation during training and, as a result, VAE collapses into a simple language model.
% Although variational inference has been
% exploited for dialogue modeling, machine translation and question
% answering~\cite{serban2016hierarchical,miao2016neural,zhang2016variational},
% so far there are no quantitative results that demonstrate that VAEs can lead to better modeling
% results than simple recurrent language models.
%
% A number of previous approaches for unsupervised sentence encoding include RNN
% language models~\cite{mikolov2010recurrent}, sequence autoencoders~\cite{dai2015semi},
% skip-thought vector~\cite{kiros2015skip} and paragraph vectors~\cite{le2014distributed}.
% The RNN~\cite{mikolov2010recurrent} represents the state of the art of language modeling.
% Sequence autoencoders~\cite{dai2015semi} have been successfully applied to
% semi-supervised classification task as a pre-training technique.
% The skip-thought model~\cite{kiros2015skip} is similar to sequence autoencoder
% but instead generates text conditioning on neighboring sentences.
% However, as pointed out by~\cite{bowman2015generating}, these models cannot
% be used in the generative setting and fail to learn a smooth and
% interpretable representation of text that contains global features
% like topic or high-level syntactic properties. This indicates the necessity of
% using probabilistic generative model for text modeling.
%
% \citet{bowman2015generating}
% uses LSTM
% %~\cite{HochreiterS97}
% as the decoder for the VAE.
% In contrast,
Related work \citep{miao2016neural,larochelle2012neural,mnih2014neural} has used simpler decoders that model text as a bag of words. Their results indicate better use of latent representations, but their decoders cannot effectively model longer-range dependencies in text and thus underperform in terms of final perplexity.

Motivated by these observations, we hypothesize that the contextual capacity of the decoder plays an important role in whether VAEs effectively condition on the latent representation when trained on text data. We propose the use of a dilated CNN as a decoder in VAE, inspired by the recent
success of using CNNs for audio, image and language modeling
~\cite{van2016wavenet, kalchbrenner2016neural, van2016conditional}. In contrast with prior work where extremely large CNNs are used, we exploit the dilated CNN for its flexibility in varying the amount of conditioning context.
In the two extremes, depending on the choice of dilation, the CNN decoder can reproduce a simple MLP using a bags of words representation of text, or can reproduce the long-range dependence of recurrent architectures (like an LSTM) by conditioning on the entire history. Thus, by choosing a dilated CNN as the decoder, we are able to conduct experiments where we vary contextual capacity, finding a sweet spot where the decoder can accurately model text but does not yet overpower the latent representation.

We demonstrate that when this trade-off is correctly managed, textual VAEs can perform substantially better than simple LSTM language models, a finding consistent with recent image modeling experiments
using variational lossy autoencoders~\cite{chen2016variational}.
We go on to show that VAEs with carefully selected CNN decoders can be quite effective for semi-supervised classification and unsupervised clustering,
outperforming several strong baselines (from ~\cite{dai2015semi}) on both text categorization and sentiment analysis.

Our contributions are as follows: First, we propose the use of a dilated CNN as a new decoder
for VAE. We then empirically evaluate several dilation architectures with different capacities, finding that
reduced contextual capacity leads to stronger reliance on latent representations.
By picking a decoder with suitable contextual capacity, we find our VAE performs better than
LSTM language models on two data sets.
% Secondly, we visualize the latent representation for these two
% data set, we find rather different types: the representation for topic falls into
% clusters while the representation for sentiment data set lies continuous on one dimension
% according to the rating.
We also explore the use of dilated CNN VAEs for semi-supervised
classification and find they perform better than strong baselines from
~\cite{dai2015semi}. Finally, we verify that the same framework can be used
effectively for unsupervised clustering.
% % Finally, we show our model can generate text conditioning on topic and sentiment label.

\section{Model}
In this section, we begin by providing background on the use of variational autoencoders for language modeling.
Then we introduce the dilated CNN architecture that we will use as a new decoder for VAE in experiments.
Finally, we describe the generalization of VAE that we will use to conduct experiments on semi-supervised classification.
% and unsupervised clustering.

\subsection{Background on Variational Autoencoders}
Neural language models~\cite{mikolov2010recurrent} typically generate each token
$x_{t}$ conditioned on the entire history of previously generated tokens:
\begin{align}
  p(\mathbf{x}) = \prod_{t}p(x_{t} | x_{1}, x_{2}, ..., x_{t-1}).
  \label{eq:lm}
\end{align}
State-of-the-art language models often parametrize these conditional probabilities
using RNNs, which compute an evolving hidden state
over the text which is used to predict each $x_{t}$. This approach, though effective in modeling text, does not explicitly model variance in higher-level properties of entire utterances (e.g. topic or style) and thus can have difficulty with heterogeneous datasets.

\citet{bowman2015generating} propose a different approach to generative text modeling inspired by related work on vision \cite{kingma2013auto}. Instead of directly modeling the joint probability $p(\mathbf{x})$
as in Equation~\ref{eq:lm}, we specify a generative process for which $p(\mathbf{x})$ is a marginal distribution. Specifically, we first generate a continuous latent vector representation $\mathbf{z}$ from a multivariate Gaussian prior $p_{\theta}(\mathbf{z})$, and then generate the text sequence $\mathbf{x}$
% is generated
% using ancestral sampling in two steps: (1) sample $\mathbf{z}$ from
% the prior distribution $p(\mathbf{z})$, (2) $\mathbf{x}$ is generated
from a conditional distribution $p_{\theta}(\mathbf{x} | \mathbf{z})$ parameterized using a neural net (often called the generation model or decoder). Because this model incorporates a latent variable that modulates the entire generation of each whole utterance, it may be better able to capture high-level sources of variation in the data. Specifically, in contrast with Equation \ref{eq:lm}, this generating distribution conditions
on latent vector representation $\mathbf{z}$:
\begin{align}
  p_{\theta}(\mathbf{x} |\mathbf{z}) = \prod_{t}p_{\theta}(x_{t} | x_{1}, x_{2}, ..., x_{t-1}, \mathbf{z}).
\end{align}

To estimate model parameters $\theta$ we
would ideally like to maximize the marginal probability
$p_{\theta}(\mathbf{x}) = \int p_{\theta}(\mathbf{z}) p_{\theta}(\mathbf{x}| \mathbf{z}) d\mathbf{z}$.
However, computing this marginal is intractable for many decoder choices. Thus, the following variational lower bound is often used as an objective \cite{kingma2013auto}:
\begin{align*}
  &\log \ p_{\theta}(\mathbf{x}) = -\log \int  p_{\theta}(\mathbf{z}) p_{\theta}(\mathbf{x}| \mathbf{z}) d\mathbf{z} \\
  % \geq &  \int q(\mathbf{z} | \mathbf{x}) \log \frac{p(\mathbf{x} |
  %        \mathbf{z})}{q(\mathbf{z} | \mathbf{z})} p(\mathbf{z})
  %        d \mathbf{z} \\
%   &\leq  \E_{q_{\phi}(\mathbf{z}|\mathbf{x})} [-\log p_{\theta}(\mathbf{x}|\mathbf{z}) - \log p_{\theta}(\mathbf{z}) +
%          \log q_{\phi}(\mathbf{z}|\mathbf{x})] \\
  \geq \ \ & \E_{q_{\phi}(\mathbf{z}|\mathbf{x})} [\log p_{\theta}(\mathbf{x}|\mathbf{z})]
       - \text{KL}(q_{\phi}(\mathbf{z}|\mathbf{x}) || p_{\theta}(\mathbf{z})).
\end{align*}
Here, $q_{\phi}(\mathbf{z} | \mathbf{x})$ is an approximation to the true posterior (often called the recognition model or encoder) and is parameterized by $\phi$. Like the decoder, we have a choice of neural architecture to parameterize the encoder. However, unlike the decoder, the choice of encoder does not change the model class -- it only changes the variational approximation used in training, which is a function of both the model parameters $\theta$ {\emph and} the approximation parameters $\phi$. Training seeks to optimize these parameters jointly using stochastic gradient ascent. A final wrinkle of the training procedure involves a stochastic approximation to the gradients of the variational objective (which is itself intractable). We omit details here, noting only that the final distribution of the posterior approximation $q_{\phi}(\mathbf{z}|\mathbf{x})$ is typically assumed to be Gaussian so that a
re-parametrization trick can be used, and refer readers to ~\cite{kingma2013auto}.

\subsection{Training Collapse with Textual VAEs}

Together, this combination of generative model and variational inference procedure are often referred to as a variational autoencoder (VAE).
% In contrast with Equation \ref{eq:lm}, this distribution conditions
% on a latent representation $\mathbf{z}$:
% \begin{align}
%   p(\mathbf{x} |\mathbf{z}) = \prod_{t}p(x_{t} | x_{1}, x_{2}, ..., x_{t-1}, \mathbf{z}).
% \end{align}
% The desired result is that learned representations $\mathbf{z}$ contains some high level information
% such as topic, which is helpful in predicting tokens
% $x_{t}$.
We can also view the VAE as a regularized version of the autoencoder. Note, however, that while VAEs are valid probabilistic models whose likelihood can be evaluated on held-out data, autoencoders are not valid models. If only the first
term of the VAE variational bound $\E_{q_{\phi}(\mathbf{z}|\mathbf{x})}[\log
p_\theta(\mathbf{x}|\mathbf{z})]$ is used as an objective,
the variance of the posterior probability $q_{\phi}(\mathbf{z}|\mathbf{x})$
will become small and the training procedure reduces to an autoencoder.
It is the KL-divergence term,
$\text{KL}(q_{\phi}(\mathbf{z}|\mathbf{x}) || p_{\theta}(\mathbf{z}))$, that discourages the VAE memorizing each $\mathbf{x}$ as a single
latent point.

While the KL term is critical for training VAEs, historically, instability on text has been evidenced by the KL term becoming vanishingly small during training, as observed by ~\citet{bowman2015generating}. When the training procedure collapses in this way, the result is an encoder that has duplicated the Gaussian prior (instead of a more interesting posterior), a decoder that completely ignores the latent variable $\mathbf{z}$, and a learned model that reduces to a simpler language model. We hypothesize that this collapse condition is related to the contextual capacity of the decoder architecture. The choice encoder and decoder depends on the type of data. For images, these are typically MLPs or CNNs. LSTMs have been used for text, but have resulted in training collapse as discussed above \cite{bowman2015generating}. Here, we propose to use a dilated CNN as the decoder instead. In one extreme, when the effective contextual width of a CNN is very large, it resembles the behavior of LSTM. When the width is very small, it behaves like a bag-of-words model. The architectural flexibility of dilated CNNs allows us to change the contextual capacity and conduct experiments to validate our hypothesis:
decoder contextual capacity and effective use of encoding information are directly related.
 We next describe the details of our decoder.

% However,
% the authors find the decoder depends too much on context information and the latent
% representation from the encoder is ignored.

% We suspect that it is the decoder model that plays an
% important role. If the decoder relies too much on context, the VAE tends to ignore the
% latent representation, turning into a standard RNN language model.
% Hence, we propose to use a dilated CNN as the decoder. The architecture
% flexibility of CNNs allows us to change the contextual capacity, hence control the
% context information and latent representation trade-off.
% In two extreme cases, when the effective contextual width of a CNN is very large, it
% resembles the behavior of LSTM and when it is very small, it behaves like a bag of words model.

\begin{figure}[!t]
  \centering
  \begin{subfigure}[t]{0.8\linewidth}
    \includegraphics[width=\linewidth]{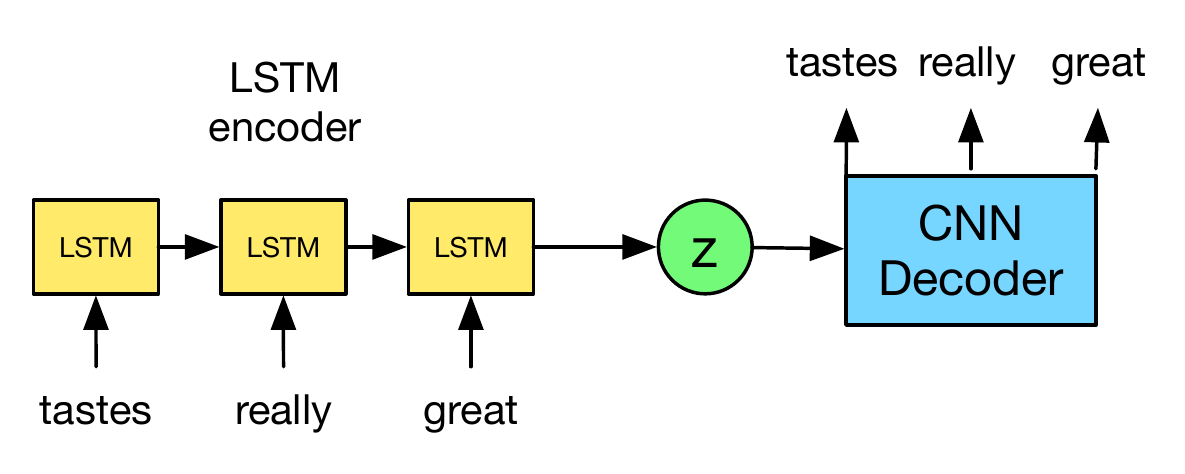}
      \caption{VAE training graph using a dilated CNN decoder.}
      \label{fig:vae}
  \end{subfigure}
  \begin{subfigure}[t]{0.8\linewidth}
      \centering
      \includegraphics[width=\linewidth]{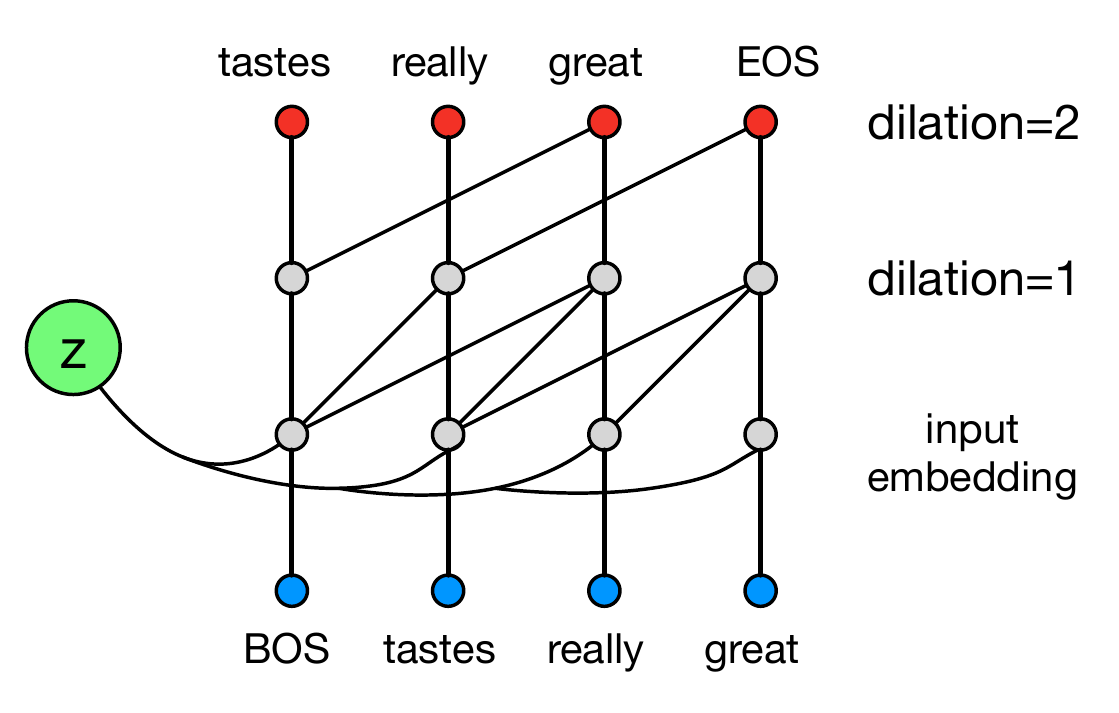}
      \caption{Digram of dilated CNN decoder.}
      \label{fig:cnn}
  \end{subfigure}
  \caption{Our training and model architectures for textual VAE using a dilated CNN decoder.}
\end{figure}

\subsection{Dilated Convolutional Decoders \label{cnn_sec}}
% CNN is less exploited for language until recently. It is used for
% language modeling and machine translation and has
% demonstrated competitive performance compared with previous state of art using
% LSTM~\cite{bahdanau2014neural}. CNN can be used as both the encoder and the decoder.
%We explore CNN as the decoder for VAE.
The typical approach to using CNNs used for text generation~\cite{kalchbrenner2016neural} is
similar to that used for images~\cite{krizhevsky2012imagenet,he2016deep},
but with the convolution applied in one dimension. We take this approach here in defining our decoder.
\\[0.2cm]
{\bf One dimensional convolution}: For a CNN to serve as a decoder for text, generation of $x_{t}$ must only condition on past tokens $x_{<t}$. Applying the
traditional convolution will break this assumption and use tokens $x_{\geq t}$ as inputs to predict
$x_{t}$. In our decoder, we avoid this by simply shifting the input by several slots~\cite{van2016conditional}. With a convolution with filter size of $k$ and using $n$ layers,
our effective filter size (the number of past tokens to condition to in
predicting $x_t$) would be $(k-1)\times n + 1$. Hence, the filter size would grow linearly with the depth of the network.
\\[0.2cm]
{\bf Dilation}: Dilated convolution~\cite{yu2015multi} was introduced to greatly increase the
effective receptive field size without increasing the computational cost. With
dilation $d$, the convolution is applied so that $d-1$ inputs are skipped each step.
Causal convolution can be seen a special case with $d=1$. With dilation, the effective
receptive size grows exponentially with network depth. In
Figure~\ref{fig:cnn}, we show dilation of sizes of 1 and 2 in the first and second layer, respectively. Suppose the dilation size in the $i$-th layer is $d_{i}$ and we use the same filter
size $k$ in all layers, then the effective filter size is $ (k-1)\sum_{i}d_{i} + 1$.
The dilations are typically set to double every layer $d_{i+1} = 2d_{i}$,
so the effective receptive field size can grow exponentially. Hence, the contextual capacity
of a CNN can be controlled across a greater range by manipulating the filter size, dilation size and network depth. We use this approach in experiments.
\\[0.2cm]
{\bf Residual connection}:
We use residual connection~\cite{he2016deep} in the decoder
\begin{wrapfigure}{r}{0.13\textwidth}
  \vspace{-0.5cm}
  \centering
  \includegraphics[width=0.13\textwidth]{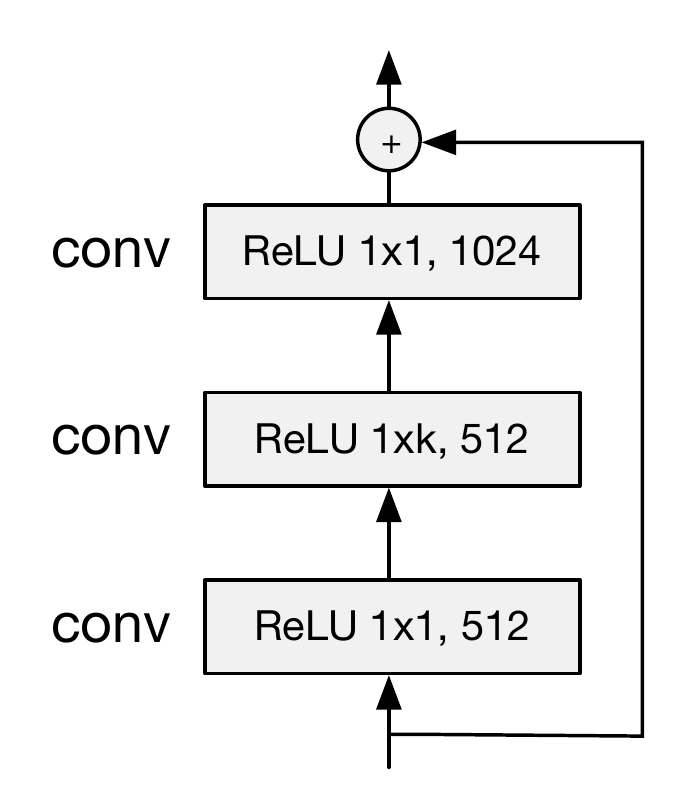}
  \label{fig:residual}
  \vspace{-0.4cm}
\end{wrapfigure}
to speed up
convergence and enable training of deeper models.
We use a residual block
(shown to the right)
similar to that
of~\cite{kalchbrenner2016neural}.
We use three convolutional layers with filter size $1\times1, 1\times k, 1\times 1$,
respectively, and ReLU activation
between convolutional layers.
% The residual block can be more powerful by
% adding batch normalization and gating
% mechanism~\cite{van2016conditional,kalchbrenner2016neural}.
\\[0.2cm]
{\bf Overall architecture}: Our VAE architecture is shown in Figure~\ref{fig:vae}.
We use LSTM as the encoder to get the posterior probability
$q(\mathbf{z}|\mathbf{x})$, which we assume to be
diagonal Gaussian. We parametrize the mean $\mathbf{\mu}$ and
variance $\mathbf{\sigma}$ with LSTM output. We sample $\mathbf{z}$ from
$q(\mathbf{z}|\mathbf{x})$, the decoder is conditioned on the sample by
concatenating $\mathbf{z}$ with every word embedding of the decoder input.

\subsection{Semi-supervised VAE}
In addition to conducting language modeling experiments, we will also conduct experiments on semi-supervised classification of text using our proposed decoder.
In this section, we briefly review
%In addition to run experiments using the VAE architecture
%described above for language modeling, we conduct experiments
semi-supervised VAEs of \cite{kingma2014semi} that incorporate discrete labels as additional variables.
Given the labeled set
$(x, y) \sim D_{L}$ and the unlabeled set
$x \sim D_{U}$, \cite{kingma2014semi} proposed a model
whose latent representation contains continuous vector $\mathbf{z}$ and discrete label
$\mathbf{y}$:
\begin{align}
  p(\mathbf{x}, \mathbf{y}, \mathbf{z}) =
  p(\mathbf{y}) p(\mathbf{z}) p(\mathbf{x}|\mathbf{y}, \mathbf{z}).
\end{align}
The semi-supervised VAE fits a discriminative network
$q(\mathbf{y} | \mathbf{x})$, an inference network
$q(\mathbf{z}| \mathbf{x}, \mathbf{y})$ and a generative network
$p(\mathbf{x}|\mathbf{y},\mathbf{z})$ jointly as part of optimizing a variational lower
bound similar that of basic VAE. For labeled data $(\mathbf{x}, \mathbf{y})$,
this bound is:
\begin{align*}
  \log p(\mathbf{x}, \mathbf{y}) \geq & \E_{q(\mathbf{z}|\mathbf{x}, \mathbf{y})} [ \log p(\mathbf{x}
                        | \mathbf{y}, \mathbf{z})] \\
                      &- \text{KL}(q(\mathbf{z}|\mathbf{x}, \mathbf{y}) ||
                        p(\mathbf{z})) + \log p(\mathbf{y}) \\
  = &  L(\mathbf{x}, \mathbf{y}) + \log p(\mathbf{y}).
\end{align*}
For unlabeled data $\mathbf{x}$, the label is treated as a latent
variable, yielding:
{\small
\begin{align*}
  \log p(\mathbf{x}) \geq & U(\mathbf{x}) \\
  = & \E_{q(\mathbf{y}|\mathbf{x})}\big[\E_{q(\mathbf{z}|\mathbf{x}, \mathbf{y})} [ \log p(\mathbf{x}
      | \mathbf{y}, \mathbf{z})]  \\
                   &- \text{KL}(q(\mathbf{z}|\mathbf{x}, \mathbf{y}) ||
                     p(\mathbf{z})) + \log p(\mathbf{y})   - \log
                     q(\mathbf{y}|\mathbf{x})\big] \\
  = & \sum_{y}q(\mathbf{y}|\mathbf{x}) L(\mathbf{x}, \mathbf{y}) -
      \text{KL}(q(\mathbf{y}|\mathbf{x}) || p(\mathbf{y})).
\end{align*}
}
Combining the labeled and unlabeled data terms, we have the overall objective
as:
\begin{align*}
  J =& \E_{(\mathbf{x}, \mathbf{y}) \sim D_{L}}[L(\mathbf{x}, \mathbf{y})] + \E_{\mathbf{x} \sim D_{U}}[U(\mathbf{x})] \\
     & + \alpha \E_{(\mathbf{x}, \mathbf{y})\sim D_{L}}[\log q(\mathbf{y}|\mathbf{x})],
\end{align*}
where $\alpha$ controls the trade off between generative and
discriminative terms.
% Since $\mathbf{y}$ is a discrete variable, we have to compute the marginal probability
% by iterating all classes. The computational cost scales linearly with the number
% of classes.
\\[0.2cm]
{\bf Gumbel-softmax}: \citet{jang2016categorical,maddison2016concrete} propose
a continuous approximation to sampling from a categorical
distribution. Let $u$ be a categorical distribution with
probabilities $\pi_{1}, \pi_{2}, ..., \pi_{c}$. Samples from $u$ can be approximated using:
\begin{align}
  y_{i} = \frac{\exp((\log (\pi_{i}) + g_{i}) / \tau)} {\sum_{j=1}^{c}
  \exp((\log(\pi_{j}) + g_{j})/\tau)},
\end{align}
where $g_{i}$ follows Gumbel(0, 1).
% We can obtain the samples from Gumbel
% distribution by first sample
% $u\sim \text{Uniform}(0, 1)$ and then compute
% $g = -\log(-\log(u))$.
The approximation is accurate when $\tau\to 0$ and smooth when $\tau > 0$.
In experiments, we use Gumbel-Softmax to approximate the samples from
$p(\mathbf{y}|\mathbf{x})$ to reduce the computational cost. As a result, we can directly
back propagate the gradients of $U(\mathbf{x})$ to the discriminator network.
We anneal $\tau$ so that sample variance is small
when training starts and then gradually decrease $\tau$.

{\bf Unsupervised clustering}:
In this section we adapt the same framework for
unsupervised clustering. We directly minimize the objective
$U(\mathbf{x})$, which is consisted of two parts: reconstruction loss and KL
regularization on $q(\mathbf{y}|\mathbf{x})$. The first part encourages the
model to assign $\mathbf{x}$ to label $\mathbf{y}$ such that the
reconstruction loss is low. We find that the model can easily get stuck in two
local optimum: the KL term is very small and $q(\mathbf{y} | \mathbf{x})$
is close to uniform distribution or the KL term is very large and all
samples collapse to one class. In order to make the model more robust,
we modify the KL term by:
\begin{align}
  \text{KL}_{\mathbf{y}} = \max(\gamma, \text{KL}(q(\mathbf{y}|\mathbf{x}) | p(\mathbf{y})).
  \label{eq:gamma}
\end{align}
That is, we only minimize the KL term when it is large enough.

\section{Experiments}
\subsection{Data sets}
\begin{table}[!bp]
\small
  \centering
  \begin{tabular}{l r r r r}
    Data & classes & documents & average \#w & vocabulary \\
    \toprule
    Yahoo & 10 & 100k & 78 & 200k \\
    Yelp15 & 5 & 100k & 96 & 90k \\
  \end{tabular}
  \caption{Data statistics}
  \label{tab:data}
\end{table}
Since we would like to investigate VAEs for language modeling and
semi-supervised classification, the data sets should be suitable for
both purposes.
% There lack existing data sets that satisfy our needs.
% The standard PTB data set~\cite{marcus1993building} used for language
% modeling can not be used for classification, while the Stanford sentiment
% classification data set~\cite{socher2013recursive} is too short and not suitable
% for language modeling. The 20news group data set~\cite{lang1995newsweeder}
% is small and only contain 10k samples.
We use two large scale document classification data sets:
Yahoo Answer and Yelp15 review, representing topic classification
and sentiment classification data sets respectively
\cite{tang2015document,yang2016hierarchical,zhang2015character}.
The original data sets contain millions of samples, of which we sample
100k as training and 10k as validation and test from
the respective partitions. The detailed statistics of both data sets are in
Table~\ref{tab:data}. Yahoo Answer contains 10 topics including Society \&
Culture, Science \& Mathematics etc. Yelp15 contains 5 level of rating, with
higher rating better.
%Our data sets are publicly available at
%\url{anonymouslink}.

\begin{table*}[!th]
  \small
  \centering
  \begin{subtable}{0.45\textwidth}
    \begin{tabular}{l r l l}
      Model & Size & NLL (KL) &  PPL \\
      \toprule
      LSTM-LM & $<i$ & 334.9 & 66.2 \\
      LSTM-VAE$^{**}$  & $<i$  & 342.1 (0.0) & 72.5 \\
      LSTM-VAE$^{**}$ + init & $<i$  & 339.2 (0.0) & 69.9 \\
      \midrule
      SCNN-LM & 15 & 345.3 & 75.5 \\
      SCNN-VAE & 15 & 337.8 (13.3) & 68.7 \\
      SCNN-VAE + init & 15 & 335.9 (13.9) & 67.0 \\
      \midrule
      MCNN-LM & 63 & 338.3 & 69.1 \\
      MCNN-VAE & 63 & 336.2 (11.8) & 67.3 \\
      MCNN-VAE + init & 63 & 334.6 (12.6) & 66.0 \\
      \midrule
      LCNN-LM & 125 & 335.4  & 66.6 \\
      LCNN-VAE & 125  & 333.9 (6.7) & 65.4 \\
      LCNN-VAE + init & 125 & {\bf 332.1 (10.0)} & {\bf 63.9} \\
      \midrule
      VLCNN-LM & 187 & 336.5  & 67.6 \\
      VLCNN-VAE & 187  & 336.5 (0.7) & 67.6 \\
      VLCNN-VAE + init & 187 & 335.8 (3.8) & 67.0 \\
\bottomrule
    \end{tabular}
    \caption{Yahoo}
    \label{tab:yahooppl}
  \end{subtable}
  \qquad
  \begin{subtable}{0.45\textwidth}
    \begin{tabular}{l r l l}
      Model & Size & NLL (KL) &  PPL \\
      \toprule
      LSTM-LM & $<i$ & 362.7 & 42.6 \\
      LSTM-VAE$^{**}$  & $<i$  & 372.2 (0.3) & 47.0 \\
      LSTM-VAE$^{**}$ + init & $<i$  & 368.9 (4.7) & 46.4 \\
      \midrule
      SCNN-LM & 15 & 371.2 & 46.6 \\
      SCNN-VAE & 15 & 365.6 (9.4) & 43.9 \\
      SCNN-VAE + init & 15 & 363.7 (10.3) & 43.1 \\
      \midrule
      MCNN-LM & 63 & 366.5 & 44.3 \\
      MCNN-VAE & 63 & 363.0 (6.9) & 42.8 \\
      MCNN-VAE + init & 63 & 360.7 (9.1) & 41.8 \\
      \midrule
      LCNN-LM & 125 & 363.5  & 43.0 \\
      LCNN-VAE & 125  & 361.9 (6.4) & 42.3 \\
      LCNN-VAE + init & 125 & {\bf 359.1 (7.6)} & {\bf 41.1} \\
      \midrule
      VLCNN-LM & 187 & 364.8 & 43.7 \\
      VLCNN-VAE & 187  & 364.3 (2.7) & 43.4 \\
      VLCNN-VAE + init & 187 & 364.7 (2.2) & 43.5 \\
      \bottomrule
    \end{tabular}
    \caption{Yelp}
    \label{tab:yelpppl}
  \end{subtable}
  \caption{Language modeling results on the test set.
  $^{**}$ is from ~\cite{bowman2015generating}.
  We report negative log likelihood (NLL) and perplexity (PPL)
  on the test set. The KL component of NLL is given in parentheses. Size indicates the
  effective filter size. VAE + init indicates pretraining of only the encoder
  using an LSTM LM.
  }
  \label{tab:ppl}
\end{table*}

\subsection{Model configurations and Training details}
We use an LSTM as an encoder for VAE and explore LSTMs and CNNs as decoders. For
CNNs, we explore several different configurations. We set the convolution filter
size to be 3 and gradually increase the depth and dilation from [1, 2, 4], [1,
2, 4, 8, 16] to [1, 2, 4, 8, 16, 1, 2, 4, 8, 16]. They represent small, medium
and large model and we name them as SCNN, MCNN and LCNN. We also explore a very large
model with dilations [1, 2, 4, 8, 16, 1, 2, 4, 8, 16, 1, 2, 4, 8, 16] and name it
as VLCNN. The effective filter size are 15, 63, 125 and 187 respectively. We
use the last hidden state of the encoder LSTM and feed it though an MLP to get the
mean and variance of $q(\mathbf{z}|\mathbf{x})$, from which we sample
$\mathbf{z}$ and then feed it through an MLP to get the starting state of decoder.
For the LSTM decoder, we follow~\cite{bowman2015generating} to use it as the initial state
of LSTM and feed it to every step of LSTM. For the CNN decoder, we concatenate it with the
word embedding of every decoder input.

The architecture of the Semi-supervised VAE basically follows that of the VAE. We
feed the last hidden state of the encoder LSTM through a two layer MLP then a
softmax to get $q(\mathbf{y}|\mathbf{x})$. We use Gumbel-softmax to
sample $\mathbf{y}$ from $q(\mathbf{y}|\mathbf{x})$. We then concatenate
$\mathbf{y}$ with the last hidden state of encoder LSTM and feed them throught an
MLP to get the mean and variance of $q(\mathbf{z}|\mathbf{y},\mathbf{x})$.
$\mathbf{y}$ and $\mathbf{z}$ together are used as the starting state of the decoder.

We use a vocabulary size of 20k for both data sets and set the word embedding
dimension to be 512. The LSTM dimension is 1024. The number of channels for
convolutions in CNN decoders is 512 internally and 1024 externally,
as shown in Section~\ref{cnn_sec}.
We select the dimension of $\mathbf{z}$ from [32,
64]. We find our model is not sensitive to this parameter.

We use Adam~\cite{kingma2014adam} to optimize all models and the learning rate is selected
from [2e-3, 1e-3, 7.5e-4] and $\beta_{1}$ is selected from [0.5, 0.9]. Empirically, we find
learning rate 1e-3 and $\beta_{1}=0.5$ to perform the best. We select drop out ratio of
LSTMs (both encoder and decoder) from [0.3, 0.5]. Following~\cite{bowman2015generating},
we also use drop word for the LSTM decoder, the drop word ratio is selected from
[0, 0.3, 0.5, 0.7]. For the CNN decoder, we use a drop out ratio of 0.1 at
each layer. We do not use drop word for CNN decoders. We use batch size of 32
and all model are trained for 40 epochs. We start to half the learning rate
every 2 epochs after epoch 30. Following~\cite{bowman2015generating}, we use KL cost
annealing strategy. We set the initial weight of KL cost term to be 0.01 and
increase it linearly until a given iteration $T$. We treat $T$ as a hyper parameter
and select it from [10k, 40k, 80k].

\subsection{Language modeling results}

\begin{figure}[!tb]
  \centering
  \includegraphics[width=0.35\textwidth]{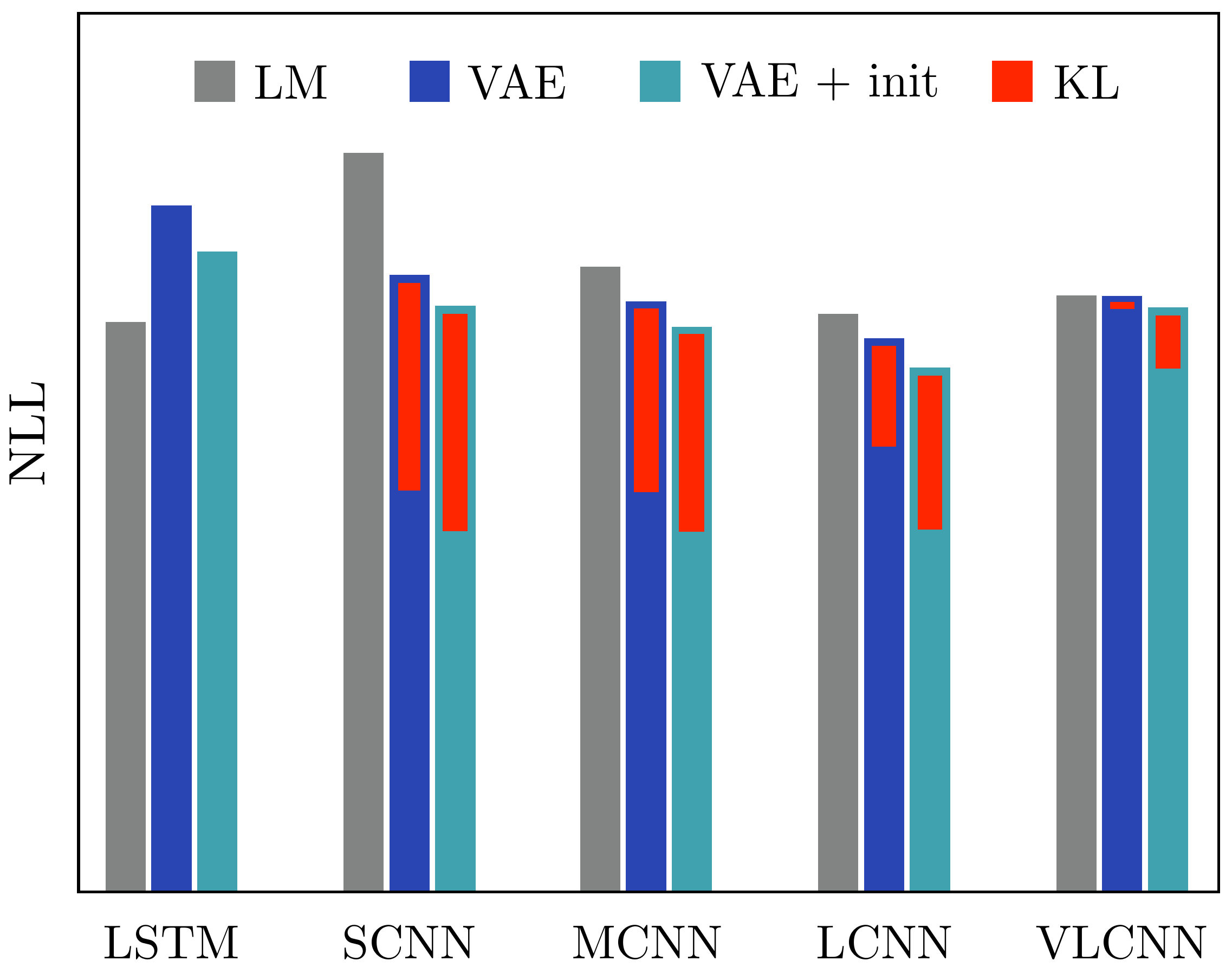}
  \caption{NLL decomposition of Table~\ref{tab:yahooppl}. Each group consists
    of three bars, representing LM, VAE and VAE+init. For VAE, we decompose the
    loss into reconstruction loss and KL divergence, shown in blue and red
    respectively. We subtract all loss values with $300$ for better
    visualization.}
  \label{fig:kl}
\end{figure}

The results for language modeling are shown in Table~\ref{tab:ppl}. We report
the negative log likelihood (NLL) and perplexity (PPL) of the test set. For the NLL of VAEs,
we decompose it into reconstruction loss and KL divergence and report the
KL divergence in the parenthesis. To better visualize these results, we plot
the results of Yahoo data set (Table~\ref{tab:yahooppl}) in Figure~\ref{fig:kl}.

We first look at the LM results for Yahoo data set. As we gradually increase
the effective filter size of CNN from SCNN, MCNN to LCNN, the NLL decreases from
345.3, 338.3 to 335.4. The NLL of LCNN-LM is very close to the NLL of LSTM-LM
334.9. But VLCNN-LM is a little bit worse than LCNN-LM, this indicates a little
bit of over-fitting.

We can see that LSTM-VAE is worse than LSTM-LM in terms of NLL and the KL
term is nearly zero, which verifies the finding
of \cite{bowman2015generating}. When we use CNNs as the decoders for VAEs,
we can see improvement over pure CNN LMs. For SCNN, MCNN and LCNN,
the VAE results improve over LM results from 345.3 to 337.8, 338.3 to 336.2, and 335.4 to 333.9
respectively. The improvement is big for small models and gradually decreases
as we increase the decoder model contextual capacity. When the model is as large as VLCNN,
the improvement diminishes and the VAE result is almost the same with LM result.
This is also reflected in the KL term, SCNN-VAE has
the largest KL of 13.3 and VLCNN-VAE has the smallest KL of 0.7.
When LCNN is used as the decoder, we obtain an optimal trade off between using
contextual information and latent representation. LCNN-VAE achieves a
NLL of 333.9, which improves over LSTM-LM with NLL of 334.9.

We find that if we initialize the parameters of {\em LSTM encoder} with
parameters of LSTM language model, we can improve the VAE results further.
This indicates better encoder model is also a key factor for VAEs
to work well. Combined with encoder initialization, LCNN-VAE improves over LSTM-LM from
334.9 to 332.1 in NLL and from 66.2 to 63.9 in PPL. Similar results for the sentiment data set are shown in Table~\ref{tab:yelpppl}. LCNN-VAE improves over LSTM-LM from 362.7
to 359.1 in NLL and from 42.6 to 41.1 in PPL.
\begin{figure}[!tb]
  \centering
  \begin{subfigure}{0.23\textwidth}
    \centering
    \includegraphics[width=\textwidth]{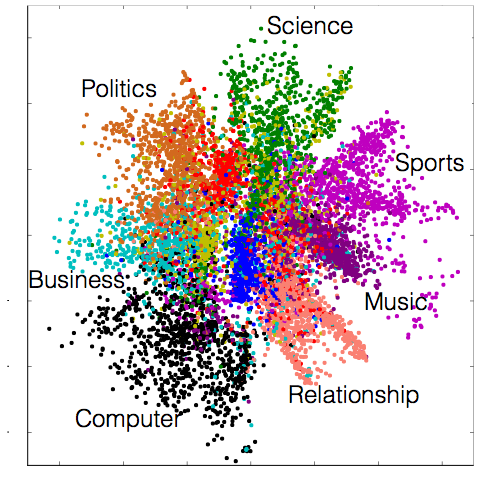}
    \caption{Yahoo}
    \label{fig:yahoo_code}
  \end{subfigure}
  \begin{subfigure}{0.23\textwidth}
    \centering
    \includegraphics[width=\textwidth]{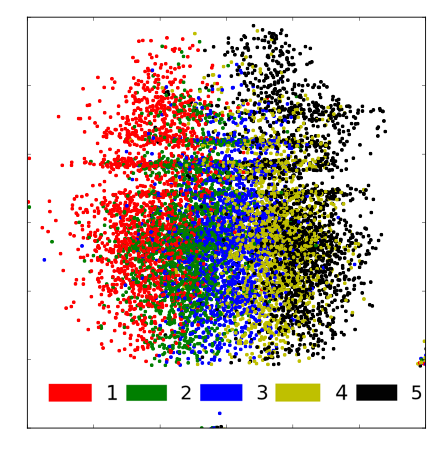}
    \caption{Yelp}
    \label{fig:yelp_code}
  \end{subfigure}
  \caption{Visualizations of learned latent representations.}
  \label{fig:code}
\end{figure}
\\[0.2cm]
{\bf Latent representation visualization:} In order to visualize the latent representation,
we set the dimension of $\mathbf{z}$ to be 2
and plot the mean of posterior probability $q(\mathbf{z}|\mathbf{x})$, as shown in
Figure~\ref{fig:code}. We can see distinct different characteristics of topic
and sentiment representation. In Figure~\ref{fig:yahoo_code}, we can see that documents
of different topics fall into different clusters, while in
Figure~\ref{fig:yelp_code}, documents of different ratings
form a continuum, they lie continuously on the x-axis as the review rating increases.
% This is consistent with sentiment actually being real-valued.

\begin{table}[!thp]
\small
  \centering
  \begin{tabular}{l r l}
    Model & ACCU & NLL (KL) \\
    \toprule
    LSTM-VAE-Semi & 51.9 & 345.5 (9.3) \\
    SCNN-VAE-Semi & {\bf 65.5} & 335.7 (10.4) \\
    MCNN-VAE-Semi & 64.6 & 332.8 (7.2) \\
    LCNN-VAE-Semi & 57.2 & {\bf 331.3} (2.7) \\
    \bottomrule
  \end{tabular}
  \caption{Semi-supervised VAE ablation results on Yahoo.
  We report both the NLL and classification
  accuracy of the test data. Accuracy is in percentage.
  Number of labeled samples is fixed to be 500.}
  \label{tab:yahoosemi_compare}
\end{table}

\begin{table*}[!thp]
\small
  \centering
  \scalebox{0.95}{
  \begin{subtable}{0.48\textwidth}
    \begin{tabular}{l r r r r}
      Model & 100 & 500 & 1000 & 2000 \\
      \toprule
      LSTM & 10.7 & 11.9 & 14.3 & 23.1 \\
      LA-LSTM~\cite{dai2015semi} & 20.8 & 42.2 & 50.4 & 54.7 \\
      LM-LSTM~\cite{dai2015semi} & 46.9 & 61.3 & 63.9 & 65.6 \\
      \midrule
      SCNN-VAE-Semi & 55.4 & 65.6 & 66.0 & 65.8\\
      SCNN-VAE-Semi+init & {\bf 63.8} & {\bf 65.4} & {\bf 66.6} & {\bf 67.4} \\
      \bottomrule
    \end{tabular}
    \caption{Yahoo}
    \label{tab:yahoosemi}
  \end{subtable}
  \quad
  \begin{subtable}{0.48\textwidth}
    \begin{tabular}{l r r r r}
      Model & 100 & 500 & 1000 & 2000 \\
      \toprule
      LSTM & 22.6 & 25.4 & 27.9 & 29.9 \\
      LA-LSTM~\cite{dai2015semi} & 35.2 & 46.4 & 49.8 & 52.2 \\
      LM-LSTM~\cite{dai2015semi} & 46.9 & 54.1 & 57.2 & 57.7 \\
      \midrule
      SCNN-VAE-Semi & 51.4 & 53.5 & 55.3 & 57.4 \\
      SCNN-VAE-Semi+init & {\bf 52.6} & {\bf 57.3} & {\bf 58.9} & {\bf 59.8} \\
      \bottomrule
    \end{tabular}
    \caption{Yelp}
    \label{tab:yelpsemi}
  \end{subtable}
  }
  \caption{Semi-supervised VAE results on the test set, in percentage.
  LA-LSTM and LM-LSTM come from ~\cite{dai2015semi}, they denotes
  the LSTM is initialized with a sequence autoencoder and a language model.}
  \label{tab:semi}
  \vspace{-0.2cm}
\end{table*}

\subsection{Semi-supervised VAE results}
Motivated by the success of VAEs for language modeling, we continue to explore
VAEs for semi-supervised learning. Following that of~\cite{kingma2014semi},
we set the number of labeled samples to be 100, 500, 1000 and 2000 respectively.
\\[0.2cm]
{\bf Ablation Study}: At first, we would like to explore the effect of
different decoders for semi-supervised classification.
We fix the number of labeled samples to be 500 and report
both classification accuracy and NLL of the test set of Yahoo data set in
Table.~\ref{tab:yahoosemi_compare}. We can see that SCNN-VAE-Semi has the best
classification accuracy of 65.5. The accuracy decreases as we gradually increase the
decoder contextual capacity. On the other hand, LCNN-VAE-Semi
has the best NLL result. This classification accuracy and NLL trade off
once again verifies our conjecture: with small contextual window size,
the decoder is forced to use the encoder information, hence the latent
representation is better learned.

Comparing the NLL results of Table~\ref{tab:yahoosemi_compare}
with that of Table~\ref{tab:yahooppl}, we can see the
NLL improves. The NLL of semi-supervised VAE improves over simple
VAE from 337.8 to 335.7 for SCNN, from 336.2 to 332.8 for MCNN,
and from 333.9 to 332.8 for LCNN.
The improvement mainly comes from the KL divergence part, this indicates that better latent
representations decrease the KL divergence, further improving
the VAE results.
\\[0.2cm]
{\bf Comparison with related methods}: We compare Semi-supervised VAE with the
methods from \cite{dai2015semi}, which represent the previous state-of-the-art for semi-supervised
sequence learning. \citet{dai2015semi} pre-trains a classifier by initializing the parameters
of a classifier with that of a language model or a sequence autoencoder. They find it improves
the classification accuracy significantly. Since SCNN-VAE-Semi performs the best
according to Table~\ref{tab:yahoosemi_compare}, we fix
the decoder to be SCNN in this part. The detailed comparison is in
Table~\ref{tab:semi}. We can see that semi-supervised VAE performs better than LM-LSTM and
LA-LSTM from~\cite{dai2015semi}. We also initialize the encoder of the VAE with parameters
from LM and find classification accuracy further improves.
We also see the advantage of SCNN-VAE-Semi
over LM-LSTM is greater when the number of labeled samples is smaller. The
advantage decreases as we increase the number of labeled samples. When we
set the number of labeled samples to be 25k, the SCNN-VAE-Semi achieves an accuracy of 70.4,
which is similar to LM-LSTM with an accuracy of 70.5. Also, SCNN-VAE-Semi performs better on Yahoo
data set than Yelp data set. For Yelp, SCNN-VAE-Semi is a little bit worse than LM-LSTM
if the number of labeled samples is greater than 100, but becomes better when we initialize the encoder.
Figure~\ref{fig:yelp_code} explains this observation. It shows the documents are coupled
together and are harder to classify. Also, the latent representation contains
information other than sentiment, which may not be useful for classification.
\subsection{Unsupervised clustering results}
\begin{table}[!thp]
  \centering
  \begin{tabular}{l r}
    Model & ACCU  \\
    \toprule
    LSTM + GMM &  25.8  \\
    SCNN-VAE + GMM  & 56.6 \\
    SCNN-VAE + init + GMM & 57.0 \\
    \midrule
    SCNN-VAE-Unsup + init & {\bf 59.9} \\
    \bottomrule
  \end{tabular}
  \caption{Unsupervised clustering results for Yahoo data set. We run each
    model 10 times and report the best results. LSTM+GMM means we extract the
    features from LSTM language model. SCNN-VAE + GMM means we use the mean of
    $q(\mathbf{z}|\mathbf{x})$ as the feature. SCNN-VAE + init + GMM means
    SCNN-VAE is trained with encoder initialization.}
  \label{tab:yahoosemi_compare}
\end{table}
% \begin{table*}[!th]
%   \centering
%   \small
%   \begin{tabular}{r  p{12cm}}
%     \toprule
%     {\bf Society} & do you think there is a god ? \\
%     {\bf Science} & how many orbitals are there in outer space ? how many orbitals are there in the solar system ? \\
%     {\bf Health} & what is the difference between \_UNK and \_UNK \\
%     {\bf Education} & what is the difference between a computer and a \_UNK ? \\
%     {\bf Computers} & how can i make flash mp3 files ? i want to know how to make a flash video so i can upload it to my mp3 player ? \\
%     {\bf Sports} & who is the best soccer player in the world ? \\
%     {\bf Business} & what is the best way to make money online ? \\
%     {\bf Music} & who is the best artist of all time ? \\
%     {\bf Relationships} & how do i know if a guy likes me ? \\
%     {\bf Politics} & what do you think about Iran ? \\
%     \bottomrule
%   \end{tabular}
%   \caption{Text generated by conditioning on topic label.}
%   \label{tab:yahoo_sample_paper}
% \end{table*}
\begin{table*}[!thbp]
  \centering
  \small
  \begin{tabular}{r  p{12cm}}
    \toprule
    {\bf 1 star} & the food was good but the service was horrible .
    took forever to get our food .
    we had to ask twice for our check after we got our food . will not return .  \\
    {\bf 2 star} & the food was good , but the service was terrible .
    took forever to get someone to take our drink order .
    had to ask 3 times to get the check . food was ok , nothing to write about . \\
    {\bf 3 star} & came here for the first time last night .
    food was good . service was a little slow . food was just ok . \\
    {\bf 4 star} & food was good , service was a little slow ,
    but the food was pretty good . i had the grilled chicken sandwich and it
    was really good . will definitely be back !
 \\
    {\bf 5 star} & food was very good , service was fast and friendly .
    food was very good as well . will be back !                     \\
    \bottomrule
  \end{tabular}
  \caption{Text generated by conditioning on sentiment label.}
  \label{tab:yelp_sample_paper}
\end{table*}
We also explored using the same framework for unsupervised
clustering. We compare with the baselines that extract the feature with
existing models and then run Gaussian Mixture Model (GMM) on these features. We
find empirically that simply using the features does not perform well since the
features are high dimensional. We run a PCA on these features, the dimension of
PCA is selected from [8, 16, 32]. Since GMM can easily get stuck in poor local
optimum, we run each model ten times and report the best result.
We find directly optimizing $U(\mathbf{x})$ does not perform well for unsupervised
clustering and we need to initialize the encoder with LSTM language model. The
model only works well for Yahoo data set. This is potentially because
Figure~\ref{fig:yelp_code} shows that sentiment latent representations does
not fall into clusters. $\gamma$ in Equation~\ref{eq:gamma} is a sensitive parameter,
we select it from the range between 0.5 and 1.5 with an interval of 0.1.
We use the following evaluation protocol~\cite{makhzani2015adversarial}: after we
finish training, for cluster $i$, we find out the validation sample $\mathbf{x}_n$ from cluster $i$
that has the best $q(y_i|\mathbf{x})$ and assign the label of $\mathbf{x}_n$ to all samples in
cluster $i$. We then compute the test accuracy based on this assignment.
The detailed results are in Table~\ref{tab:yahoosemi_compare}. We can see
SCNN-VAE-Unsup + init performs better than other baselines. LSTM+GMM performs very
bad probably because the feature dimension is 1024 and is too high for
GMM, even though we already used PCA to reduce the dimension.
\\[0.2cm]
{\bf Conditional text generation}
With the semi-supervised VAE, we are able to generate text conditional on the
label. Due to space limitation, we only show one example of generated reviews
conditioning on review rating in Table~\ref{tab:yelp_sample_paper}.
% More examples of text
% generated conditioning on topic and rating are shown in the Appendix.
For
each group of generated text, we fix $\mathbf{z}$ and vary the label $\mathbf{y}$, while picking $\mathbf{x}$ via beam search with a beam size of 10.

\section{Related work}
Variational inference via the re-parameterization trick was initially proposed
by~\cite{kingma2013auto, rezende2014stochastic} and since then, VAE has been
widely adopted as generative model for images~\cite{gregor2015draw,
  yan2016attribute2image, salimans2015markov, gregor2016towards, hu2017unifying}.

Our work is in line with previous works on combining variational inferences
with text
modeling~\cite{bowman2015generating,miao2016neural,serban2016hierarchical,
  zhang2016variational, hu2017controllable}. \cite{bowman2015generating} is the first work to
combine VAE with language model and they use LSTM as the decoder and find some
negative results. On the other hand, \cite{miao2016neural} models text as bag
of words, though improvement has been found, the model can not be used to
generate text.  Our work fills the gaps between
them. \cite{serban2016hierarchical, zhang2016variational} applies variational
inference to dialogue modeling and machine translation and found some
improvement in terms of generated text quality, but no language modeling
results are
reported. \cite{chung2015recurrent,bayer2014learning,fraccaro2016sequential}
embedded variational units in every step of a RNN, which is different from our model
in using global latent variables to learn high level features.

Our use of CNN as decoder is inspired by recent success of PixelCNN model for
images~\cite{van2016conditional}, WaveNet for audios~\cite{van2016wavenet},
Video Pixel Network for video modeling~\cite{kalchbrenner2016video} and ByteNet
for machine translation~\cite{kalchbrenner2016neural}. But in contrast to those
works showing using a very deep architecture leads to better performance, CNN
as decoder is used in our model to control the contextual capacity, leading to better performance.

Our work is closed related the recently proposed variational lossy
autoencoder~\cite{chen2016variational} which is used to predict image
pixels. They find that conditioning on a smaller window of a pixels leads to better
results with VAE, which is similar to our finding.
Much~\cite{rezende2015variational, kingma2016improving, chen2016variational} has been done to
come up more powerful prior/posterior distribution representations with techniques such as
normalizing flows. We treat this as one of our future works.
This work is largely orthogonal and could be potentially
combined with a more effective choice of decoder to yield additional gains.

There is much previous work exploring unsupervised sentence encodings, for example
skip-thought vectors~\cite{kiros2015skip}, paragraph
vectors~\cite{le2014distributed}, and sequence
autoencoders~\cite{dai2015semi}. \cite{dai2015semi} applies a pretrained
model to semi-supervised classification and find significant gains, we use this
as the baseline for our semi-supervised VAE.

\section{Conclusion}

We showed that by controlling the decoder's contextual capacity in VAE, we can improve performance on both language modeling and semi-supervised classification tasks by preventing a degenerate collapse of the training procedure. These results indicate that more carefully characterizing decoder capacity and understanding how it relates to common variational training procedures may represent important avenues for unlocking future unsupervised problems.

% We propose to use dilated CNNs as decoders for VAEs for text modeling. We studied the
% contextual information and latent representation trade off by varying the
% decoder contextual capacity through changing CNN
% architectures. We find with a decoder with a small context window, the VAE is forced to use
% information from the latent representation. By selecting a suitable decoder, the VAE can
% perform better than simple LSTM language models. We find a similar trade off
% between classification accuracy and NLL for semi-supervsied VAEs.
% We show our semi-supervised VAEs perform better than strong baselines with proper decoders
% are selected. There are several future directions to explore based on our work.
% The first is to use more sophisticated prior/posterior probability representations
% such as inverse autoregressive flow to further improve the VAE results. Anther direction is to
% come up with better models for sentiment analysis with VAE since it has shown
% rather different code structure with topic.

\bibliography{bible}
\bibliographystyle{icml2017}

\appendix
\begin{table*}[!thbp]
  \centering
  \small
  \begin{tabular}{r  p{12cm}}
    \toprule
    {\bf Society} & do you think there is a god ? \\
    {\bf Science} & how many orbitals are there in outer space ? how many orbitals are there in the solar system ? \\
    {\bf Health} & what is the difference between \_UNK and \_UNK \\
    {\bf Education} & what is the difference between a computer and a \_UNK ? \\
    {\bf Computers} & how can i make flash mp3 files ? i want to know how to make a flash video so i can upload it to my mp3 player ? \\
    {\bf Sports} & who is the best soccer player in the world ? \\
    {\bf Business} & what is the best way to make money online ? \\
    {\bf Music} & who is the best artist of all time ? \\
    {\bf Relationships} & how do i know if a guy likes me ? \\
    {\bf Politics} & what do you think about Iran ? \\
    \midrule
    {\bf Society} & what is the meaning of life ? \\
    {\bf Science} & what is the difference between kinetic energy and heat ? \\
    {\bf Health} & what is the best way to get rid of migraine headaches ? \\
    {\bf Education} & what is the best way to study for a good future ? \\
    {\bf Computers} & what is the best way to install windows xp home edition ? \\
    {\bf Sports} & who do you think will win the super bowl this year ? \\
    {\bf Business} & i would like to know what is the best way to get a good paying
                     job ? \\
    {\bf Entertainment} & what do you think is the best movie ever ? \\
    {\bf Relationships} & what is the best way to get over a broken heart ? \\
    {\bf Politics} & what do you think about the war in iraq ? \\
    \midrule
    {\bf Society} & what would you do if you had a million dollars ? \\
    {\bf Mathematics} & i need help with this math problem ! \\
    {\bf Health} & what is the best way to lose weight ? \\
    {\bf Education} & what is the best college in the world ? \\
    {\bf Computers} & what is the best way to get a new computer ? \\
    {\bf Sports} & who should i start ? \\
    {\bf Business} & what is the best way to get a good paying job ? \\
    {\bf Entertainment} & who do you think is the hottest guy in the world ? \\
    {\bf Relationships} & what should i do ? \\
    {\bf Politics} & who do you think will be the next president of the united states ? \\
    \midrule
    {\bf Society} & do you believe in ghosts ? \\
    {\bf Science} & why is the sky blue ? \\
    {\bf Health} & what is the best way to get rid of a cold ? \\
    {\bf Reference} & what do you do when you are bored ? \\
    {\bf Computers} & why ca n't i watch videos on my computer ? when i try to watch
                      videos on my computer , i ca n't get it to work on my computer
                      . can anyone help ? \\
    {\bf Sports} & what do you think about the \_UNK game ? \\
    {\bf Business} & what is the best way to get a job ? \\
    {\bf Entertainment} & what is your favorite tv show ? \\
    {\bf Relationships} & how do you know when a guy likes you ? \\
    {\bf Politics} & what do you think about this ? \\
    \midrule
    {\bf Society} & what is the name of the prophet muhammad ( pbuh ) ? i do n't know if
                    he is a jew or not . \\
    {\bf Science} & where can i find a picture of the \_UNK \_UNK \_UNK \_UNK ? i need to know
                    the name of the insect that has the name of the whale . \\
    {\bf Health} & what is the best way to get rid of a \_UNK mole ?  \\
    {\bf Reference} & does anyone know where i can find info on \_UNK \_UNK \_UNK ? i am
                      looking for the name of the \_UNK \_UNK . \\
    {\bf Computers} & does anyone know where i can find a picture of a friend 's cell
                      phone ? \\
    {\bf Sports} & does anyone know where i can find a biography of \_UNK ? \\
    {\bf Business} & does anyone know where i can find a copy of the \_UNK ? \\
    {\bf Music} & does anyone know the name of the song and who sings it ? \\
    {\bf Relationship} & how do i tell my boyfriend that i love him ? he is my best
                         friend , but i dont know how to tell him . please help ! ! !
                         ! ! ! \\
    {\bf Politics} & where is osama bin laden ? \\
    \bottomrule
  \end{tabular}
  \caption{Text generated by conditioning on topic label.}
  \label{tab:yahoo_sample}
\end{table*}

\begin{table*}[!thbp]
  \centering
  \small
  \begin{tabular}{r  p{12cm}}
    \toprule
    {\bf 1 star} & the food is good , but the service is terrible . i have been
                   here three times and each time the service has been horrible
                   . the last time we were there , we had to wait a long time
                   for our food to come out . when we finally got our food ,
                   the food was cold and the service was terrible . i will not
                   be back . \\
    {\bf 2 star} & this place used to be one of my favorite places to eat in
                    the area . \\
    {\bf 3 star} & i 've been here a few times , and the food has always been
                    good . \\
    {\bf 4 star} & this is one of my favorite places to eat in the phoenix
                    area . the food is good , and the service is friendly . \\
    {\bf 5 star} & my husband and i love this place . the food is great , the
                    service is great , and the prices are reasonable . \\
    \midrule

    {\bf 1 star} & this is the worst hotel i have ever been to . the room was dirty
                   , the bathroom was dirty , and the room was filthy .\\
    {\bf 2 star} & my husband and i decided to try this place because we had heard
                    good things about it so we decided to give it a try . the
                    service was good , but the food was mediocre at best . \\
    {\bf 3 star} & we came here on a saturday night with a group of friends . we
                    were seated right away and the service was great . the food was
                    good , but not great . the service was good and the atmosphere
                    was nice . \\
    {\bf 4 star} & my husband and i came here for brunch on a saturday night . the
                    place was packed so we were able to sit outside on the patio
                    . we had a great view of the bellagio fountains and had a great
                    view of the bellagio fountains . we sat at the bar and had a
                    great view of the bellagio fountains . \\
    {\bf 5 star} & my husband and i came here for the first time last night and
                    had a great time ! the food was amazing , the service was great
                    , and the atmosphere was perfect . we will be back ! \\
    \midrule

    {\bf 1 star} & this is the worst place i have ever been to . i will never go
                   back . \\
    {\bf 2 star} & i was very disappointed with the quality of the food and the
                   service . i will not be returning . \\
    {\bf 3 star} & this was my first time at this location and i have to say it
                   was a good experience . \\
    {\bf 4 star} & this is a great place to grab a bite to eat with friends or
                   family . \\
    {\bf 5 star} & i am so happy to have found a great place to get my nails done
                   . \\
    \midrule
    {\bf 1 star} & my wife and i have been going to this restaurant for years
                   . the last few times i have been , the service has been
                   terrible . the last time we were there , we had to wait a
                   long time for our food to arrive . the food is good , but
                   not worth the wait . \\
    {\bf 2 star} & the food is good , but the service leaves something to be
                   desired . \\
    {\bf 3 star} & i have been here a few times . the food is consistently
                   good , and the service is good . \\
    {\bf 4 star} & my wife and i have been here a few times . the food is
                   consistently good , and the service is friendly . \\
    {\bf 5 star} & my husband and i have been coming here for years . the food
                   is consistently good and the service is always great . \\
    \midrule
    {\bf 1 star} & the food was good but the service was terrible . we had to
                   wait 45 minutes for our food to come out and it was cold . i
                   will not be back .\\
    {\bf 2 star} & the food was good but the service was
                   terrible . we had a party of 6 and the food
                   took forever to come out . the food was good
                   but not worth the price . \\
    {\bf 3 star} & the food was good but the service was a little slow . we
                   had to wait a while for our food and it was n't even busy
                   . \\
    {\bf 4 star} & i have been here a few times and have never been
                   disappointed . the food was great and the service was great
                   . we will be back . \\
    {\bf 5 star} & my husband and i have been here a few times and have never
                   been disappointed . the food was great and the service was
                   great . i will definitely be back ! \\
    \midrule
    {\bf 1 star} & if i could give this place zero stars i would . i do not
                   recommend this place to anyone ! \\
    {\bf 2 star} & i do n't know what all the hype is about this place , but i
                   do n't think i will be back . \\
    {\bf 3 star} & i do n't know what all the hype is about this place , but i
                   do n't think i 'll be back .\\
    {\bf 4 star} & i 've been here a couple of times and have never been
                   disappointed . the food is fresh , the service is friendly
                   , and the prices are reasonable . \\
    {\bf 5 star} & this is the best ramen i 've ever had in my life , and i 've
                   never had a bad meal here ! \\
    \midrule
    {\bf 1 star} & this is the worst company i have ever dealt with . they do
                   n't know what they are doing . \\
    {\bf 2 star} & this is the worst buffet i have ever been to in my life
                   . the food was just ok , nothing to write home about . \\
    {\bf 3 star} & not a bad place to stay if you 're looking for a cheap
                   place to stay . \\
    {\bf 4 star} & this is a great place to stay if you 're looking for a
                   quick bite . \\
    {\bf 5 star} & i love this place ! the staff is very friendly and helpful
                   and the price is right ! \\
    \bottomrule
  \end{tabular}
  \caption{Text generated by conditioning on sentiment label.}
  \label{tab:yelp_sample}
\end{table*}

\end{document}